\documentclass[10pt,twocolumn,letterpaper]{article}

\usepackage[pagenumbers]{cvpr} % To force page numbers, e.g. for an arXiv version

\usepackage{graphicx}
\usepackage{amsmath}
\usepackage{amssymb}
\usepackage{booktabs}

\usepackage{algorithm}
\usepackage{algorithmicx}
\usepackage{algpseudocode}

\usepackage[pagebackref,breaklinks,colorlinks]{hyperref}

\usepackage[capitalize]{cleveref}
\crefname{section}{Sec.}{Secs.}
\Crefname{section}{Section}{Sections}
\Crefname{table}{Table}{Tables}
\crefname{table}{Tab.}{Tabs.}

\begin{document}

\title{Unsupervised Visual Defect Detection with
Score-Based Generative Model}

% \author{
% Yapeng Teng$^1$\\
% \and
% Haoyang Li$^1$\\
% \and
% Fuzhen Cai$^1$\\
% \and
% Ming Shao$^2$\\
% \and
% Siyu Xia$^1$

% \and
\author{Yapeng Teng$^1$
\and
Haoyang Li$^1$
\and
Fuzhen Cai$^1$
\and
Ming Shao$^2$
\and
Siyu Xia$^1$
\and
{\small $^1$ Southeast University\ \ \ \ $^2$ University of Massachusetts Dartmouth}
% \and
}
\maketitle

%%%%%%%%% ABSTRACT
\begin{abstract}
  Anomaly Detection (AD), as a critical problem, has been widely discussed. In this paper, we specialize in one specific problem, Visual Defect Detection (VDD), in many industrial applications. And in practice, defect image samples are very rare and difficult to collect. Thus, we focus on the unsupervised visual defect detection and localization tasks and propose a novel framework based on the recent score-based generative models, which synthesize the real image by iterative denoising through stochastic differential equations (SDEs). Our work is inspired by the fact that with noise injected into the original image, the defects may be changed into normal cases in the denoising process (i.e., reconstruction). First, based on the assumption that the anomalous data lie in the low probability density region of the normal data distribution, we explain a common phenomenon that occurs when reconstruction-based approaches are applied to VDD: normal pixels also change during the reconstruction process. Second, due to the differences in normal pixels between the reconstructed and original images, a time-dependent gradient value (i.e., score) of normal data distribution is utilized as a metric, rather than reconstruction loss, to gauge the defects. Third, a novel $T$ scales approach is developed to dramatically reduce the required number of iterations, accelerating the inference process. These practices allow our model to generalize VDD in an unsupervised manner while maintaining reasonably good performance. We evaluate our method on several datasets to demonstrate its effectiveness.
\end{abstract}

\begin{figure}[t]
  \centering
  \includegraphics[width=1.0\linewidth]{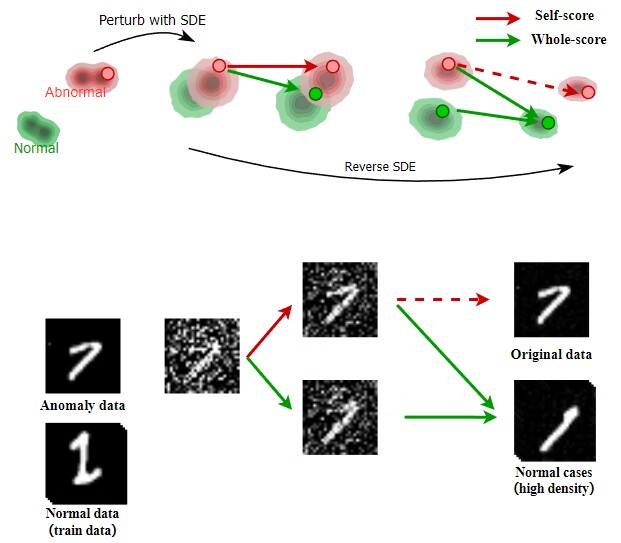}

  \caption{The red and green contour plots depict the distributions of normal and abnormal data, respectively; the red dot represents an abnormal sample, and the green dot represents a being-reconstructed sample that is gradually transforming into normal cases. Given an anomaly data, on the left, the noise is injected to blur it. When the model is iterated through \textbf{whole-score}, the defect is recovered into normal modes, as indicated by the green arrow, while through \textbf{self-score}, the noise image is restored to the original image, as indicated by the red arrow. Two whole-score values pointing to the normal cases indicate that the upper noise image, whose defect is gradually back to its original, requires a larger \textbf{whole-score} value than the lower image, which has been changing defect, to correct the defect.}
  \label{Idea}
\end{figure}
%%%%%%%%% BODY TEXT
\section{Introduction}
\label{sec:intro}

AD, whose objective is to detect previously unseen rare objects or events, plays key roles in a variety of applications, including industrial manufacturing~\cite{PaulBergmann2019MVTecA,VitjanZavrtanik2021DRAEMA,JinleiHou2021DivideandAssembleLB,ChunLiangLi2021CutPasteSL,reiss2021panda} and medical analysis~\cite{schlegl2017unsupervised,schlegl2019f,ouyang2020self,tang2021recurrent}. VDD issues, a specific sub-problem of AD, which we define in this paper as where the visual variation in the known classes is generally modest, such as unified backgrounds, and defects manifest as localized patches with anomalous visual appearances to prevent ambiguity, have been valued in various industrial products. As anomalous samples are rare in real-world scenarios, this poses challenges, especially for supervised learning models. Alternative solutions through generative models trained only on normal samples in an unsupervised manner have been prevailing recently, including Auto-encoder (AE)~\cite{doi:10.1126/science.1127647}, Generative Adversarial Network (GAN)~\cite{goodfellow2014generative}, Flow~\cite{dinh2016density} and their variants. Nonetheless, difficulty remains in applying these methods to high-dimensional data such as images. For example, AE is known for its blurred reconstructions and indistinguishable defects, and GAN or Flow models need additional overhead in developing encoders or dedicated dimensionality reduction modules, which is both time- and computational-consuming.

To effectively and directly apply generative models to VDD, we propose \textbf{Score-DD}, which is characterized by leveraging the recent score-based generative modeling through SDEs~\cite{YangSong2020ScoreBasedGM} and employing the score (i.e., the gradient of the log probability density with respect to data) instead of conventional reconstruction loss as a metric. Generally, similar to diffusion models~\cite{JonathanHo2020DenoisingDP}, score-based generative models gradually convert a pure Gaussian noise vector into a similar realistic training image through iterating an equation containing a score value, which we renamed ``\textbf{whole-score}'' in this paper and can also be used to measure the distance from the high probability density region of the training data. Furthermore, we design another iteration process by \textbf{self-score}, which guides samples back to the original. The key idea of Score-DD is shown in~\cref{Idea}. Given a defect image, we first add some noise to it, and then start to stimulate two distinct processes using whole-score and self-score. In detail, with iteration through whole-score, the noise data is gradually brought closer to the high density region of the normal data distribution by the guidance of whole-score; on the other hand, based on the easy-to-meet assumption in the VDD setting that all anomalous data exist in the low density region of the normal data distribution, the noise images obtained by iteration through self-score will gradually reveal defects far from the high density region. The full iterative processes are not required; notably, after getting two noise images in the middle, we can calculate their whole-score values to evaluate their distances from the high density region, and their divergence will be leveraged for defect detection and localization. As shown in the middle of \cref{Idea}, the upper noise image ``7'' is further away from the high density region than the lower one.

Our main contribution is to solve three challenges. First, our observations have identified and explained key issues that are common phenomena when reconstruction-based methods are applied to VDD: the reconstructed normal pixels do not exactly match the original image; therefore, a simple pixel-wise comparison between the reconstructed and original data for VDD is not reliable. Second, based on the initial assumption, we propose a new metric through the whole-score, instead of traditional reconstruction loss, to mitigate this issue. Third, the reverse process of score-model is less-efficient for certain setting of hyperparameters, e.g., a larger initial moment $t$ in our case. Instead of launching a large $t$, we propose to investigate a set of smaller parameters, i.e., $\{t\}$ with only a few steps in reverse for each. This ensemble strategy allows us to consider different ``reconstructed and original'' data pairs and enables a more reliable detection mechanism, termed $T$ scales. 

Besides, most existing methods rely on pre-trained networks for feature extraction, external data, extra mechanisms for good performance, or they need to redesign their model structure and loss function. Our goal is to explore the characteristics of the score model applied to unsupervised VDD and provide a simple and effective scheme not dependent on other models or mechanisms. Unlike them, our score model is just trained on normal data in an unsupervised fashion. And it can also easily be extended and combined with other mechanisms because our method does not modify the training procedure of the score-based generative model. We evaluate our method on several datasets to verify its effectiveness. Among them, our method achieves the state-of-the-art (SOTA) \textbf{98.24 image-level AUC} and \textbf{97.78 pixel-level AUC} on the challenging MVTec AD dataset~\cite{PaulBergmann2019MVTecA}.

\begin{figure*}[t]
  \centering
  
  \includegraphics[width=0.95\linewidth]{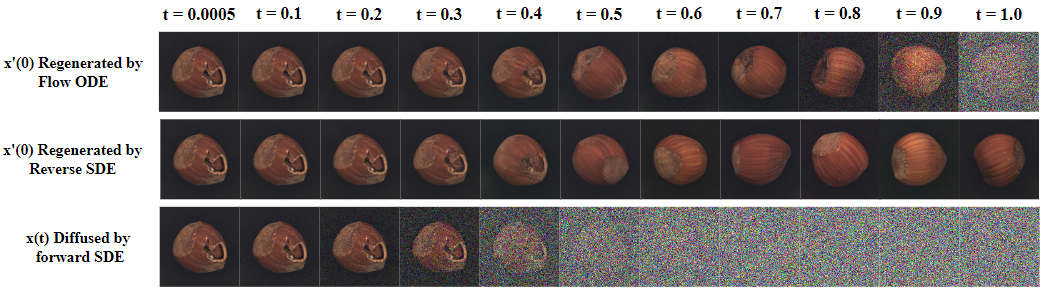}

  \caption{Regenerate samples by solving the probability flow ODE and reverse SDE on $[0 , t]$ with initial point $\mathbf{x}(t)=\mu(t)\mathbf{x}(0)+\sigma (t)\mathbf{z}(t)$ for $\mathbf{z}(t)\sim \mathcal{N}(0,\textup{I})$ trained on MVTec AD dataset.}
  \label{reconstruction}
\end{figure*}

\section{Related work}

In this section, we mainly review previous approaches based on generative models. AE is trained to generate normal data but fails to reconstruct the abnormal samples. However, the output is often blurred~\cite{JinleiHou2021DivideandAssembleLB}, or defects are well restored~\cite{VitjanZavrtanik2021ReconstructionBI} due to the nature of generalization. To fix these problems, recent works have developed and discussed the memory mechanism~\cite{gong2019memorizing,JinleiHou2021DivideandAssembleLB}, SSIM Loss~\cite{wang2004image}, Mask strategy~\cite{VitjanZavrtanik2021ReconstructionBI}, denoising autoencoder~\cite{ChaoqinHuang2019AttributeRF} and forgery defect~\cite{VitjanZavrtanik2021DRAEMA}. However, these methods have recently been superseded by the following competitive generative models.

Recently, GAN and its generative and discriminative networks have been leveraged in AD tasks. In detail, the generative networks learn to map the noise from a latent space to anomaly-free data distribution, while the discriminant network determines whether it comes from anomaly-free data distribution. However, as GAN lacks dedicated encoders to produce hidden variables of the input data, additional efforts are required to develop networks to search for the hidden variables~\cite{schlegl2019f,akccay2019skip,akcay2018ganomaly}. 

Another branch of work is based on the approach ``normalized flow'' which learns and manages to map the distribution of normal data reversely to a simple Gaussian distribution. The distribution of normal data supposes to be close to the center of the Gaussian kernels (i.e., high-density region), while the abnormal data shall exist in the low density region, an indicator for data with defects in the testing phase. However, as the hidden layer dimension must match the data dimension in these methods, when working on data of a larger size, e.g., high-resolution images, the model parameters expand quickly. Therefore, the flow-based methods~\cite{MarcoRudolph2020SameSB,MarcoRudolph2022FullyCC,DenisAGudovskiy2021CFLOWADRU,JiaweiYu12022FastFlowUA} usually take feature maps extracted by a network pre-trained on a large-scale dataset, e.g., ImageNet.

\section{Background}

\subsection{Score-based Generative Model for VDD}
\label{sec3.1}
% By converting noise data into ``normal'' ones, DAE enables to localize defects by comparing the reconstructions with the original inputs. 

The unsupervised VDD model discussed in this paper is trained on a normal dataset $\mathbf{X}_N$ in an unsupervised fashion while tested on a blend of a normal and abnormal dataset $\mathbf{X}_{N+A}$, where the main objects of the abnormal data contain defect patches. Our framework is inspired by the Denoising AutoEncoder (DAE)~\cite{ChaoqinHuang2019AttributeRF}. We can extend the DAE by integrating the diffusion process with score-based generative models \cite{ChenlinMeng2021SDEditIS,JongminYoon2021AdversarialPW}. In particular, we can model a diffusion process by using forward SDE to inject a certain amount of noise into the data and implement the reverse diffusion process as denoising. Ideally, the defects would be treated as noise and be recovered to normal data. This procedure is shown in \cref{reconstruction}. With the reconstructed data gained from the reverse diffusion, we are able to compare it with the original data through certain metrics to detect defects. We will briefly introduce: (1) diffusion process; (2) reverse process; (3) the probability flow ODE; (4) whole-score and self-score .
% in the followings.

% We base all methods on the continuous time diffusion generative model through stochastic calculus~\cite{YangSong2020ScoreBasedGM}. In this section, we first introduce basic concepts of stochastic differential equation, and background of score-based generative model.

\textbf{Diffusion process.} Diffusion process gradually adds noise to the original data $\mathbf{x}$ through forward SDE \cite{YangSong2020ScoreBasedGM}, and yields a sequence $\{\mathbf{x}(t)\}$ through:
\begin{equation}\tag{1}
\mathrm{d\mathbf{x}}(t) = f(t)\mathbf{x}(t)\mathrm{d}t+g(t)\mathrm{d\mathbf{w}}(t),\label{con:SDE}
\end{equation}
where $t\in[0,1]$ indicates the time stamp, $\mathbf{w}(t)$ denotes a standard Wiener process, and the drift coefficient $f(t)$ and the diffusion coefficient $g(t)$ are fixed. Therefore, it is essentially an ordinary differential equation (ODE) driven by the noise. We can interpret $\mathrm{d\mathbf{w}}(t)$ as an infinitesimal Gaussian noise. The solution to this diffusion process in \cref{con:SDE} is $\{\mathbf{x}(t)\}_{t\in[0,1]}$. Assume $p_{t}(\mathbf{x})$ denotes the probability density of solution and $p_{0t}(\mathbf{x}(t)|\mathbf{x}(0))$ denotes the transition distribution from $\mathbf{x}(0)$ to $\mathbf{x}(t)$. By definition, $p_{data}(\mathbf{x})\approx p_{0}(\mathbf{x})$. Based on \cref{con:SDE}, we can continuously add noises to the original data $\mathbf{x}(0) \sim p_0(\mathbf{x})$. This process gradually removes details and structure of the data as $t$ increases, and the distribution of noise data $p_{1}(\mathbf{x})$ satisfies a tractable prior distribution $\pi(\mathbf{x})$. 

% The three different forms of SDEs presented in~\cite{YangSong2020ScoreBasedGM} all satisfy the above process.

\textbf{Reverse process.} Diffusion process starts from $\mathbf{x}(0)$ and ends up with $\mathbf{x}(t)$. The reverse process aims to recover the original data from $\mathbf{x}(t)$ and get an similar value, $\mathbf{x}'(0)$, generated by the reverse of a diffusion process of \cref{con:SDE}, which is also a diffusion process and can be achieved by:
\begin{equation}\tag{2}
\mathrm{d\mathbf{x}}(t) = (f(t)\mathbf{x}(t)-g(t)^{2}\nabla_{\mathbf{x}}\mathrm{log}\ p_{t}(\mathbf{x}(t)))\mathrm{d}\bar{t}+g(t)\mathrm{d\mathbf{\bar{w}}}(t),\label{con:RSDE}
\end{equation}
where $\mathrm{d}\bar{t}$ represents negative time step, $\mathbf{\bar{w}}$ denotes a standard Wiener process in the reversal time direction. Therefore, the objective of score-based generative model transforms to learn the score function $\nabla_{\mathbf{x}}\mathrm{log}\ p_{t}(\mathbf{x}(t))$ in \cref{con:RSDE}. We can estimate $\nabla_{\mathrm{\mathbf{x}}}\mathrm{log}\ p_{t}(\mathrm{\mathbf{x}})$ by training a score-based model $s_{\theta }(\mathbf{x}(t),t)$ on 
training dataset $\mathbf{X}_N$, where $s_{\theta }(\mathbf{x}(t),t)$ adopts a variant of U-net that requires both $\mathbf{x}(t)$ and $t$ inputs. The goal now is to minimize the following loss \cite{PascalVincent2011ACB}:
\begin{equation}\tag{3}
\begin{split}
\mathcal{L}(\theta;\lambda (\cdot )):=\frac{1}{2}\int_{0}^{1}\mathbb E_{p_{0}(\mathbf{x})p_{0t}(\mathbf{x}(t)|\mathbf{x}(0))}[\lambda(t)||\\
\nabla_{\mathbf{x}}\mathrm{log}\ p_{0t}(\mathbf{x}(t)|\mathbf{x}(0))-s_{\theta }(\mathbf{x}(t),t)||_{2}^{2}]\mathrm{d}t,\label{Loss:DSM}
\end{split}
\end{equation}
which is equivalent to a constant that is irrelevant to $\theta$. Additionally, if the drift coefficient $f(t)$ is linear, the $p_{0t}(\mathbf{x}(t)|\mathbf{x}(0))=\mathcal{N}(\mathbf{x}(t);\mu(t)\mathbf{x}(0),\sigma^{2}(t)\textup{I})$ will be a tractable Gaussian distribution, and 
\begin{equation}\tag{4}
\mathbf{x}(t)=\mu(t)\mathbf{x}(0)+\sigma (t)\mathbf{z}(t),\label{transition distribution}
\end{equation}
where $\mathbf{z}(t)\sim \mathcal{N}(0,\textup{I})$. Fortunately Variance Exploding (VE), Variance Preserving (VP) and sub-VP SDE introduced in \cite{YangSong2020ScoreBasedGM} satisfy the linear drift coefficient condition (check more details in Appendix A.3), and therefore, $\nabla_{\mathbf{x}}\mathrm{log}\ p_{0t}(\mathbf{x}(t)|\mathbf{x}(0)) = -\frac{\mathbf{z}(t)}{\sigma(t)}$ of each sample can be solved. Following this, we are able to train a score-based model $s_{\theta }(\mathbf{x}(t),t)$ by sampling $\mathbf{x}(0)\sim p_0(\mathbf{x})$ from training dataset, uniformly and randomly sampling $t$ in [0,1], and getting $\mathbf{x}(t)\sim  p_{0t}(\mathbf{x}(t)|\mathbf{x}(0))$.

% After training, we can get a score-based model that $s_{\theta }(\mathbf{x},t)\approx\nabla_{\mathbf{x}}\mathrm{log}\ p_{t}(\mathbf{x})$ for almost all time and $\mathbf{x}$ in dataset. Just replacing $\nabla_{\mathbf{x}}\mathrm{log}\ p_{t}(\mathbf{x}(t))$ with $s_{\theta }(\mathbf{x}(t),t)$ in Eq.(\ref{con:RSDE}), we can start from $\mathbf{x}(1)\sim p_1(\mathbf{x})$ to obtain $\mathbf{x}(0)$ by solving Eq.(\ref{con:RSDE}) step by step. Actually, when training score-based model, we are sampling continuously to make $s_{\theta}(\mathbf{x}(t),t)\approx \nabla_{\mathbf{x}}\mathrm{log}\ p_{t}(\mathbf{x})\approx\frac{1}{n}\sum_{d=1}^{n}\nabla_{\mathbf{x}}\mathrm{log}\ p_{t}(\mathbf{x}(t)|\mathbf{x}_{d}(0))$ where the approximate equality holds when $n \to \infty $ and $n$ is number of data samples. 
\begin{figure*}
  \centering
  \begin{subfigure}[t]{\linewidth}
    \includegraphics[width=0.9\linewidth]{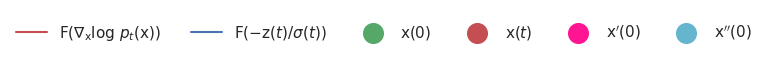}
  \end{subfigure}
  \begin{subfigure}[t]{\linewidth}
  \centering
  \begin{subfigure}{0.3\linewidth}
    \includegraphics[width=0.95\linewidth]{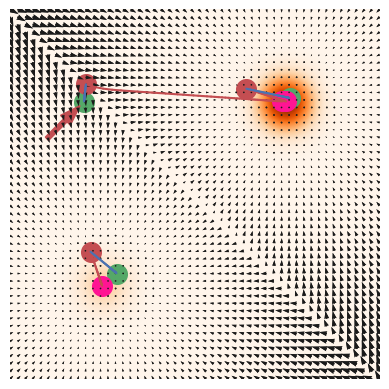}
    \caption{Flow ODE ~$t$ = 0.6}
    \label{a}
  \end{subfigure}
  \begin{subfigure}{0.3\linewidth}
    \includegraphics[width=0.95\linewidth]{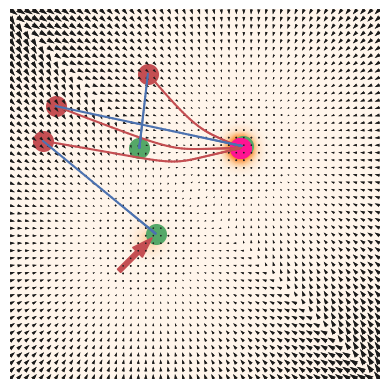}
    \caption{Flow ODE ~$t$ = 1.0}
    \label{b}
  \end{subfigure}
  \begin{subfigure}{0.3\linewidth}
    \includegraphics[width=0.95\linewidth]{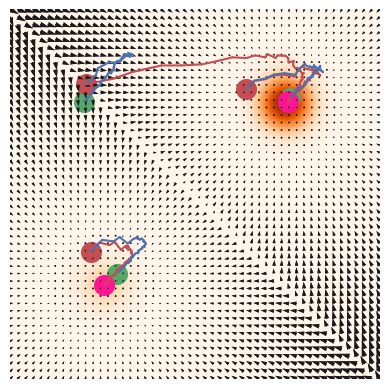}
    \caption{Reverse SDE ~$t$ = 0.6}
    \label{c}
  \end{subfigure}
  \end{subfigure}
 
  \caption{Exploratory experiment based on the VE SDE. $p_{data}(\mathbf{x})$ is shown in an orange colormap. The red trajectory represents the reverse diffusion process from $\mathbf{x}(t)$ to $\mathbf{{x}'}(0)$ driven by $\nabla_{\mathbf{x}}\mathrm{log}\ p_{t}(\mathbf{x}(t))$, and the blue trajectory is the process from $\mathbf{x}(t)$ to $\mathbf{{x}''}(0)$ driven by $-\mathbf{z}(t)/\sigma(t)$. $(a)$ Iterating the Flow ODE with $t=0.6$. $(b)$ Iterating the Flow ODE with $t=1.0$. $(c)$ Iterating the Reverse SDE with $t=0.6$. With arrows, we label the abnormal data $\mathbf{{x}}(0)$ in $(a)$, and the normal data $\mathbf{{x}}(0)$ in $(b)$ whose corresponding $\mathbf{{x}'}(0)$ eventually moves to other high-density region.}
  \label{EExp}
\end{figure*}

\textbf{Probability flow ODE}. In addition to the \cref{con:RSDE}, there is an alternative solution to the reverse diffusion process termed probability flow ODE~\cite{DimitraMaoutsa2020InteractingPS,YangSong2020ScoreBasedGM}, abbreviated as Flow ODE. The Flow ODE shares the same marginal distribution $p_t(\mathbf{x})$ with SDE of \cref{con:SDE} and can be defined as:
\begin{equation}\tag{5}
\mathrm{d}\mathbf{x}(t)=(f(t)\mathbf{x}(t)-\frac{1}{2}g(t)^{2}\nabla_{\mathbf{x}}\mathrm{log}\ p_{t}(\mathbf{x}(t)))\mathrm{d}t.\label{PFODE}
\end{equation}
In particular, the Flow ODE does not include random terms but retains the same score function $\nabla_{\mathbf{x}}\mathrm{log}\ p_{t}(\mathbf{x}(t))$. Therefore, we can also plug $s_{\theta}(\mathbf{x}(t),t)$ into \cref{con:RSDE} or \cref{PFODE} to generate samples. 

\textbf{Whole-score and self-score.}
It can be seen that the output of the trained score model $s_{\theta}(\mathbf{x}, t) \approx \nabla_{\mathbf{x}}\mathrm{log}\ p_{t}(\mathbf{x}(t))$ is the gradient pointing to the high-density regions of $\mathbf{X}_N$, and plays a key role in the reverse process. Therefore, we term it as \textbf{whole-score} $s_w(\mathbf{x}, t)$. In addition, we use score $s_e(\mathbf{x}, t) = -\frac{\mathbf{z}(t)}{\sigma(t)}$ relatively to each sample denoted as \textbf{self-score}. Both whole- and self-score will be discussed and used in our proposed Score-DD model.

\begin{figure*}[t]
    \centering
    \includegraphics[scale = 0.8]{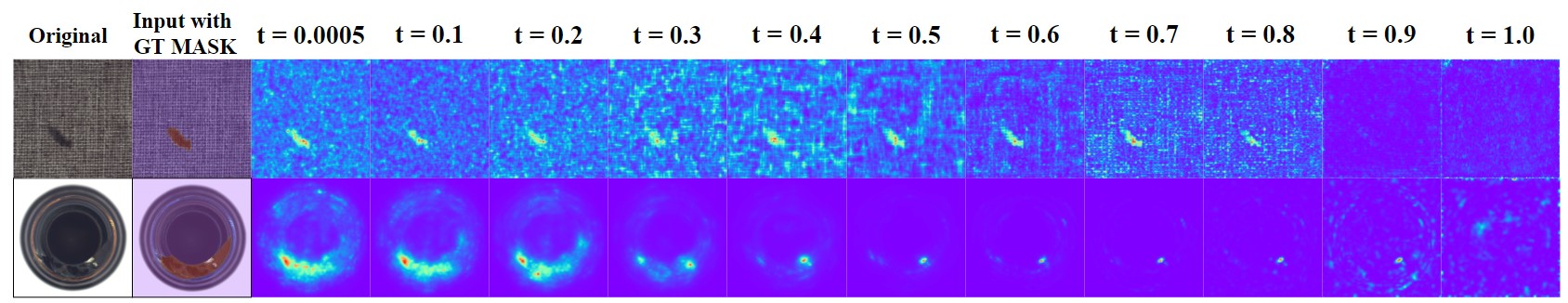}
    \caption{Different semantic information of score model at different moments. Set different initial time $t$, iterate $r=1$ steps with the Flow ODE, and present upsampling $64\times64$ feature map of score model.}
    \label{T Map}
\end{figure*}

\subsection{Issues and Observations}
\label{problems}

% Actually, there is another score $\nabla_{{x}''}logp_{0t}({x}''|x(0)) = \sum_{d=1}^{n=1}logp_{t}({x}''|x(0)^{d}) = -\frac{z(t)}{\sigma(t)}$ we discussed above, which is calculated based on each sample, which we name it as \textbf{self-score} to prevent confusion, while the trained model eventually gets the score value based on the whole data set, then renamed \textbf{whole-score} in our paper. Besides the direction of \textbf{self-score} value points to the original data $x(0)$. Specially, if we know the \textbf{self-score} value at each time $t$, with the Flow ODE, we can get the original $x(0)$ after iteration.

\label{sectionC}

% Following the diffusion and reverse processes,
We regenerate a series of images as shown in \cref{reconstruction}. In particular, given a test image, with predetermined $t\in(0,1]$, we inject noise into $\mathbf{x}(0)$ according to the forward SDEs to achieve $\mathbf{x}(t)$. Simulating reverse process by reverse SDE or Flow ODE, we are allowed to reconstruct images $\mathbf{x}'(0)$. Note that with different $t\in\{0.0005, 0.1,...\}$, details and structures are gradually removed from left to right in the third row of \cref{reconstruction}. Based on the difference between the original and reconstructed images, we can localize the defects. For example, if using the reconstruction of $t=0.5$ in the first row of \cref{reconstruction}, we may easily locate the defects. However, picking an appropriate value of $t$ is not trivial. When using a smaller $t$, not all defects can be well changed into the normal mode. On the other hand, when using a larger $t$, the normal pixels in the reconstructed image are slightly different from the original image in pixel space.

To provide more insights behind these phenomena, we conduct another experiment in \cref{EExp} to explore how the score-based model transforms anomalous data into a normal pattern and the reasons for deterioration in normal regions of the original images. For demonstration purposes, we consider VE SDE where $\mathrm{d}\mathbf{x}(t)=\sqrt{\frac{\mathrm{d}[\sigma ^{2}(t)]}{\mathrm{d}t}}\mathrm{d}\mathbf{w}(t)$. Assume that the distribution of positive data is $\frac{1}{5}\mathcal{N}((-5,-5),\textup{I})+\frac{4}{5}\mathcal{N}((5,5),\textup{I})$, and set 100 time steps in [0,1]. Other details are presented in {Appendix B}. Following the steps discussed above, we set the initial diffusion time step $t$ and obtain $\mathbf{x}(t)$ by \cref{transition distribution}, and then conduct the reverse process through Flow ODE in \cref{PFODE} or reverse SDE in \cref{con:RSDE}. In \cref{EExp}, anomalous data points originally located in low probability density move to high probability density regions driven by whole-score $s_w(\mathbf{x}, t)$ and end up as $\mathbf{x}'(0)$. For normal data, as shown in \cref{a} and \cref{c}, with a suitable value of $t$, e.g., $t=0.6$, the deviation of the $\mathbf{{x}'}(0)$ from $\mathbf{x}(0)$ is smaller. When $t=1.0$, some normal data marked by arrows in \cref{b} has trouble returning to the vicinity of $\mathbf{x}(0)$. The primary reason is that $s_w(\mathbf{x}, t)$ is learned to enforces the $\mathbf{x}'(0)$ moving towards the high probability density region of training data. When $t=1.0$, mixed Gaussians are fused into a tractable Gaussian distribution $p_1(\mathbf{x})\approx\pi(\mathbf{x})$. Therefore, some normal data are driven into the other Gaussian cores, which deviate significantly from $\mathbf{x}(0)$.

\section{Proposed Method}
% \label{sectionC}

% \noindent Therefore, in this section, we will introduce the main components of the method, and we will provide detailed principle explanation and experimental demonstration to prove the feasibility of this method.

\subsection{Leveraging Whole-scores to Localize Defects}

One of the major characters identified in \cref{EExp} is that the whole-score drives the noise data towards the high density regions, which makes anomalous data, originally located in the low density region, eventually falls into the high density regions; unexpectedly, this also makes it difficult to return to the vicinity of the original data for some normal data. Thus, we shall carefully select $t$ value to retain the overall contour of the distribution, such that the reconstructed normal data remain in the vicinity of the original data after iterations. Otherwise, the normal region of the original data will be changed significantly, as shown in \cref{reconstruction} when $t>0.5$. This stringent requirement on $t$ makes comparing the difference between the reconstructed and original image in pixel space a less feasible or reliable solution. However, we find that normal and abnormal data behave differently in the dimension of probability density. From \cref{EExp}, after sufficient iterations, all normal or abnormal data eventually fall into the high density portion of the normal data distribution. Based on the assumption that original anomaly data is in the low probability density zone and normal data is originally in the high density region, we propose that employing a metric connected to the probability density of normal data, e.g., whole-score $s_w(\cdot,t)\approx\nabla_{\mathbf{x}}\mathrm{log}\ p_{t}(\mathbf{x}(t))$, is effective for VDD. As a result, we can feed $\mathbf{x}'(0)$ and $\mathbf{x}(0)$ into score model to assess their whole-score difference.

% Through the above analysis, an important point is that, the whole-score drives the noise data towards the high probability density region, which makes anomalous data, originally located in the low probability density region, eventually fall into the high probability density region. We also need to pick a suitable initial time point $t$ to retain the overall contour of the distribution, such that the reconstructed normal data remain in the vicinity of the original data after iteration, because of whole-score without information of $\mathbf{x}(0)$. This is an extremely stringent requirement because, in actuality, the distribution is extremely complex and uncertain, and there is no way to precisely determine an initial time $t$ that meets the aforementioned criteria.

% The key to success is to retain a good representation for the original data, especially for the normal region, and here we refer to the self-score, i.e., the gradient pointing to the original data. Intuitively, for abnormal data, there will be a large divergence between the self-score $s_e$ and the whole-score $s_w$. On the other hand, for normal data, this divergence is small. Therefore, we can use $\|s_e-s_w\|_\ell$ as the basic guidance for Score-DD modeling.

\begin{figure*}[t]
    \centering
    \includegraphics[scale = 0.56]{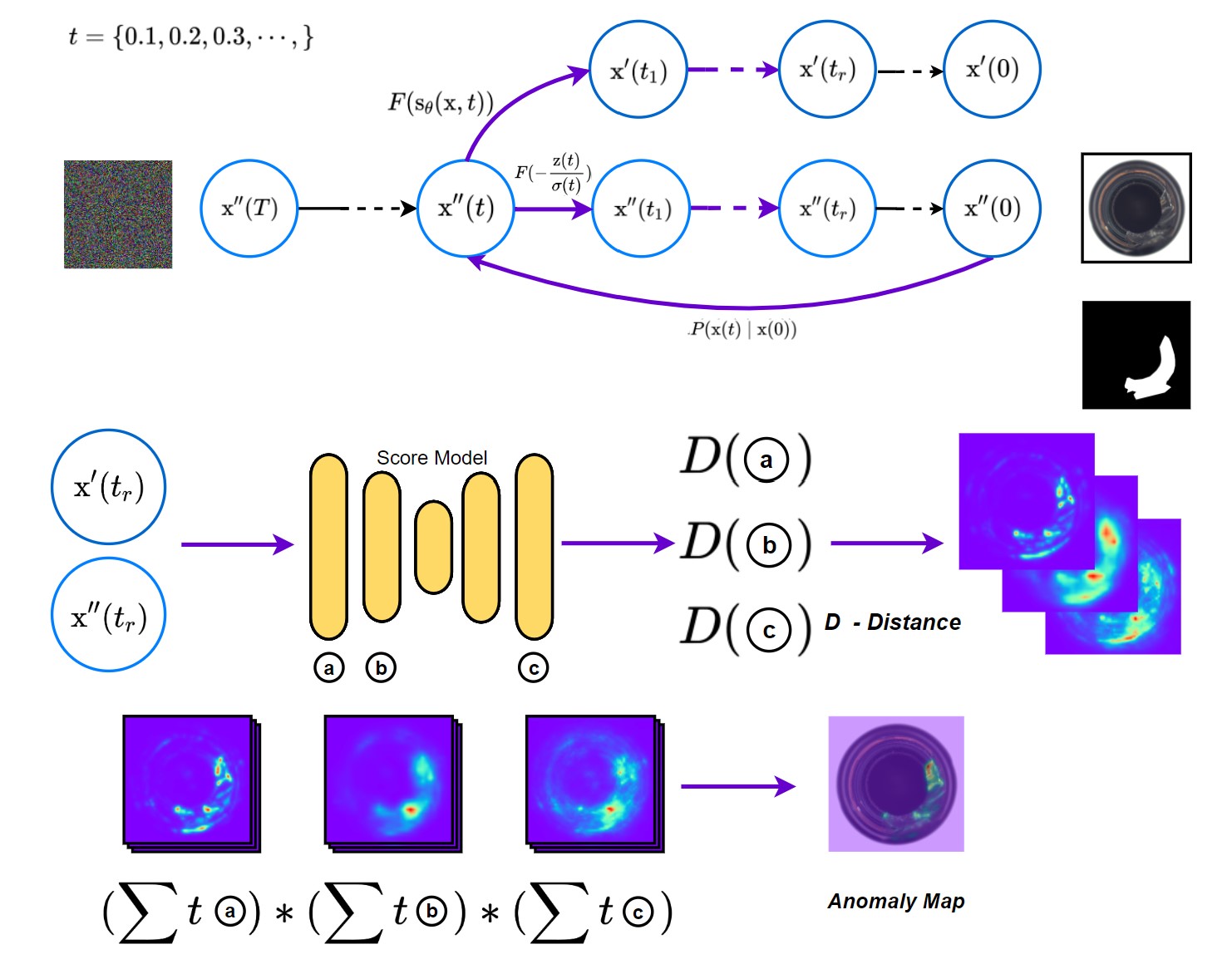}
    \caption{An overview of \textbf{Score-DD} for VDD. Set $\{t\}$ and in each moment $t$ case, first inject noise to a test image through $p_{0t}(\mathbf{x}(t)|\mathbf{x}(0))$ \cref{transition distribution}. Then solving \cref{con:RSDE} or \cref{PFODE} by plugging $s_w(\mathbf{x},t)$ and $s_e(\mathbf{x},t)$ separately into them. After iterating $r$ steps, input two samples into score model to extract feature maps. Add up all feature maps of the same resolution at different $\{t\}$, then after upsampling, the final anomaly map is obtained by multiplying them up.}
    \label{FlowChart}
    %\vspace{-0.4cm}
\end{figure*}

% While we may use the difference in probability density as a metric for AD based on assumption that original anomaly data is in the low probability density zone and anomaly-free data is in the high probability density region. Naturally, the output of score-based model, $\nabla_{\mathbf{x}}\mathrm{log}\ p_{t}(\mathbf{x}(t))$ is ideal for this task. Furthermore, the score model requires simultaneous input of $\mathbf{{x}}(t)$ and $t$. When $t$ is very small, $p_{0}(\mathbf{x})\approx p_{data}(\mathbf{x})$, by definition. Alternatively, we can input original data, reconstructed data and smallest $t$ to score-based model altogether. As a result, we lessen the impact of the problem that the reconstructed normal image differs in pixel space from the original normal image by using whole-score value as an AD metric.

\subsection{Enhancement through Feature Maps}

As the score-based model is usually implemented through neural networks, the characteristics of scores can also be reflected by feature maps. Feature maps in deep layers contain more semantic information, while shallow layers are capable of identifying fine-grained information such as lines and colors. Previous works have already investigated the usefulness of feature maps in different network layers for unsupervised AD or VDD~\cite{wang2021student,yamada2021reconstruction,JieYang2020DFRDF} as well as semantic segmentation~\cite{DmitryBaranchuk2022LabelEfficientSS} through the middle layers of a score-based model. Therefore, we also adopt the multi-scale feature maps to strengthen the performance through the U-net architecture as the following.

To avoid using more models, we adopt a similar scheme to process feature maps in \cite{yamada2021reconstruction}. Firstly, calculate the Euclidean distance between the two feature maps after performing $l_{2}$ normalization on each feature map. Then, feature maps of the same resolution are summed up, and all feature maps are scaled up to the same resolution by using the ``bilinear'' interpolations. The products of all feature maps are taken as the output. In our practice, however, we found that the $l_{2}$ normalization compromised the feature maps' efficacy and their visual effects. We believe the reason is that feature maps in a score-based model are not particularly trained for VDD in an supervised manner as in \cite{yamada2021reconstruction}. However, without normalization, feature maps of the same resolution may have different magnitudes. When they are added together, the output will favor the feature maps with greater magnitude, leading to poor results. Therefore, we skip both $l_{2}$ normalization and the sum of feature maps of the same resolution. Instead, after we directly calculate the Euclidean distance, feature maps with significant visual effects are selected, and their point-wise products will be used as outputs, as shown in \cref{FlowChart}.

\subsection{$T$ Scales}

Another issue identified in practice is the time spent in the reverse process given a large $t$ value. It takes many iterations with score model to return to $\mathbf{x}'(0)$. We are considering whether we can leverage the feature maps of the score model at different moments without the full iteration since the goal is to detect defects rather than generate images.

%In addition, a large $t$ may drive $\mathbf{x}(t)$ far from the $\mathbf{x}(0)$, and therefore the difference of whole-score value between $\mathbf{x}'(0)$ and $\mathbf{x}(0)$ are not very distinguishable as expected in some extreme cases.

% Because to get the final reconstruction need many iterations steps, especially construct a score-based model directly on large image, the speed is too slow to apply in practical application. Furthermore, while using the score value as an AD metric can help solve the problem of reconstructed normal data differing from the original data in numerical space, you can use any extreme example to show that when reconstructed normal data deviates too far from the original data, the difference in score value may no longer be distinguishable from abnormal data.

It has been discussed that the score-based model provides semantics at different moments $t$~\cite{DmitryBaranchuk2022LabelEfficientSS}, as shown in \cref{T Map}. Because the feature maps are changing gradually, there may be a lot of redundant information in the feature maps at adjacent moments. Therefore, we are motivated to not perform a full iteration to get the final image, but to iterate just $r$ steps ($r$ is a very small integer), and then compute the difference of $s_w(\cdot,t_r)$ between $\mathbf{x}'(t_r)$ and $\mathbf{x}''(t_r)$ to be the representative semantic information in a certain time period around $t$, where $\mathbf{x}''(t)$ represents the true trajectory from $\mathbf{x}(t)$ to $\mathbf{x}(0)$. In order to leverage different information at different moments, we can apply a set of different moments $\{t\}$ of capacity $T$, and in each $t$ case, we do the same process as above. We term this approach as \textbf{$T$ scales}, because it will yield $T$ feature maps to be assembled for anomaly map, as shown in \cref{FlowChart}.

Assume that at step $t$, we will examine $\mathbf{x}$ at $t_1>...>t_i>...>t_r$ in a sequential manner, where $t_i-t_{i+1}=\Delta t$ will be used as the approximation of $\textup{d}t$ in ODE. Without loss of generality, we elaborate the process of computing $\mathbf{x}'(t_i)$ and $\mathbf{x}''(t_i)$ in each $t$ case as follows. First, we can achieve $\mathbf{x}'(t_i)$ by replacing $\nabla_{\mathbf{x}}\mathrm{log}\ p_{t}(\mathbf{x}(t))$ in~\cref{con:RSDE} or~\cref{PFODE} with whole-score $s_w(\cdot,t)$ to build the reverse path as the red path in~\cref{EExp}. Second, we can model $\mathbf{x}''(t_i)$ in adjacent steps by replacing the $\nabla_{\mathbf{x}}\mathrm{log}\ p_{t}(\mathbf{x}(t))$ in~\cref{con:RSDE} or~\cref{PFODE} with self-score $s_e(\cdot,t)$ as the true trajectory from $\mathbf{x}(t)$ to $\mathbf{x}(0)$. (The proof process can be found in {Appendix A.1}). Therefore, if we know self-score $s_e(\cdot,t)$ relative to $\mathbf{x}(0)$ at each moment, we can approach the original $\mathbf{x}(0)$ from $\mathbf{x}(t)$, as the blue path in~\cref{EExp}. We elaborate the overall framework of our algorithm in~\cref{FlowChart}. Specific calculation processes and pseudo-code are given in {Appendix A.2}.

\begin{figure*}[t]
    \centering
    \includegraphics[scale = 0.6]{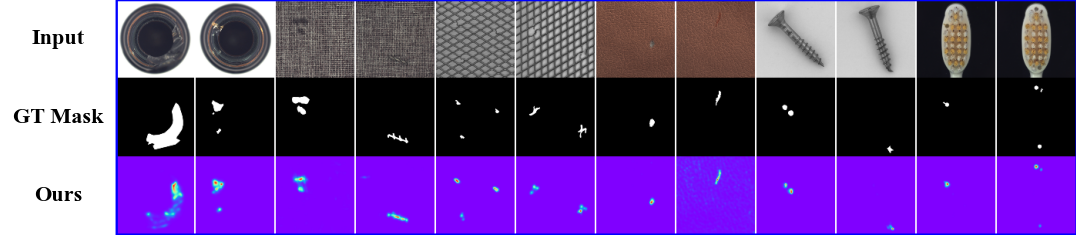}
    \setlength{\abovecaptionskip}{0.cm}
    \caption{Defect location examples of \textbf{Score-DD} on MVTec. The first line of the images is the original anomaly images. The second line is GT MASK. The third line is anomaly map from Score-DD. }
    \label{Output maps}
\end{figure*}

\begin{table*}[t]\normalsize
\centering
\setlength{\tabcolsep}{3pt}
\scalebox{0.8}{
\begin{tabular}{c|cccccccc|ccc} 
\hline
Method           & RIAD                       & OCR-GAN         & CFlow$^\star $                       & FastFlow$^\star $  & PaDiM$^\star $     & PatchCore$^\star $ & CutPaste                    & DRÆM                        & Ours$^{\romannumeral1}$                        & Ours$^{\romannumeral2}$      & Ours$^{\romannumeral3}$       \\ 
\hline
carpet           & 84.2/96.3                  & 99.4/-         & \textbf{100}/\textbf{99.3}  & \textbf{100}/99.4           & -/99.1          & 98.7/98.9          & 93.9/98.3                   & 97.0/95.5                   & 96.9/98.9                  & 86.0/95.5                   & 93.0/98.0           \\
grid             & 99.6/98.8                  & 99.6/-         & 97.6/99.0                   & 99.7/98.3                   & -/97.3          & 98.2/98.7          & \textbf{100}/97.5           & 99.9/\textbf{99.7}          & \textbf{100}/\textbf{99.7} & 100/99.6                    & 100/99.5            \\
leather          & \textbf{100}/99.4          & 97.1/-         & 97.7/\textbf{99.7}          & \textbf{100}/\textbf{99.5}  & -/99.2          & \textbf{100}/99.3  & \textbf{100}/\textbf{99.5}  & \textbf{100}/98.6           & 99.6/99.3                  & 97.8/97.9                   & 98.3/99.0           \\
tile             & 98.7/89.1                  & 95.5/-         & \textbf{98.7}/\textbf{98.0} & \textbf{100}/96.3           & -/94.1          & 98.7/95.6          & 94.6/90.5                   & \textbf{99.6}/\textbf{99.2} & 98.6/94.4                  & 98.7/95.7                   & 100/94.8            \\
wood             & 93.0/85.8                  & 95.7/-         & \textbf{99.6}/\textbf{96.7} & \textbf{100}/\textbf{97.0}  & -/94.9          & 99.2/95.0          & 99.1/95.5                   & 99.1/\textbf{96.4}          & 98.8/95.1                  & 98.9/96.9                   & 96.2/95.5           \\ 
\hline
bottle           & 99.9/98.4                  & 99.6/-         & \textbf{100}/\textbf{99.0}  & \textbf{100}/97.7           & -/98.3          & \textbf{100}/98.6  & 98.2/97.6                   & 99.2/\textbf{99.1}          & \textbf{100}/97.9          & \textbf{100}/98.1           & 99.7/95.7           \\
cable            & 81.9/84.2                  & 99.1/-         & \textbf{100}/97.6           & \textbf{100}/\textbf{98.4}  & -/96.7          & 99.5/\textbf{98.4} & 81.2/90.0                   & 91.8/94.7                   & 96.8/97.5                  & 95.7/95.1                   & 98.0/97.9           \\
capsule          & 88.4/92.8                  & 96.2/-         & \textbf{99.3}/\textbf{99.0} & \textbf{100}/\textbf{99.1}  & -/98.5          & 98.1/98.8          & 98.2/97.4                   & 98.5/94.3                   & 96.1/98.6                  & 91.5/97.4                   & 93.2/97.7           \\
hazelnut         & 83.3/96.1                  & 98.5/-         & 96.8/98.9                   & \textbf{100}/99.1           & -/98.2          & \textbf{100}/98.7  & 98.3/97.3                   & \textbf{100}/\textbf{99.7}  & 99.9/99.2                  & 98.4/\textbf{99.4}          & 97.6/99.0           \\
metal Nut        & 88.5/92.5                  & 99.5/-         & 91.9/\textbf{98.6}          & \textbf{100}/98.5           & -/97.2          & \textbf{100}/98.4  & 99.9/93.1                   & 98.7/\textbf{99.5}          & 97.2/97.9                  & 98.9/98.0                   & 99.1/94.7           \\
pill             & 83.8/95.7                  & 98.3/-         & \textbf{99.9}/\textbf{99.0} & \textbf{99.4}/\textbf{99.2} & -/95.7          & 96.7/97.1          & 94.9/95.7                   & 98.9/97.6                   & 95.3/96.0                  & 88.3/96.4                   & 93.7/94.4           \\
screw            & 84.5/98.8                  & \textbf{100}/- & \textbf{99.7}/\textbf{98.9} & 97.8/99.4                   & -/98.5          & 98.1/99.4          & 88.7/96.7                   & 93.9/97.6                   & 99.6/\textbf{99.6}         & 98.3/99.8                   & 99.1/99.8           \\
toothbrush       & \textbf{100}/\textbf{98.9} & 98.7/-         & 95.2/\textbf{99.0}          & 94.4/\textbf{98.9}          & -/98.8          & \textbf{100}/98.7  & 99.4/98.1                   & \textbf{100}/98.1           & 99.8/98.3                  & 98.3/97.8                   & 98.9/97.9           \\
transistor       & 90.9/87.7                  & 98.3/-         & 99.1/\textbf{98.0}          & \textbf{99.8}/\textbf{97.3} & -/\textbf{97.5} & \textbf{100}/96.3  & 96.1/93.0                   & 93.1/90.9                   & 95.4/95.2                  & 95.4/94.2                   & 96.4/94.7           \\
zipper           & 98.1/97.8                  & 99.0/-         & 98.5/99.1                   & 99.5/98.7                   & -/98.5          & 98.8/98.8          & \textbf{99.9}/\textbf{99.3} & \textbf{100}/98.8           & 99.8/\textbf{99.3}         & \textbf{99.9}/\textbf{99.3} & \textbf{99.9}/99.2  \\ 
\hline
\textit{Average} & 91.7/94.2                  & 98.3/-         & 98.3/\textbf{98.6}          & \textbf{99.4}/\textbf{98.5} & 97.9/97.5       & \textbf{99.1}/98.1 & 96.1/96.0                   & 98.0/97.3                   & 98.2/\textbf{97.8}         & 96.4/97.4                   & 97.5/97.2           \\
\hline
\end{tabular}}
\caption{Defect detection (\textbf{left}) and localization (\textbf{right}) performance on the MVTec AD dataset. Methods achieved for the top two AUROC (\%) are highlighted in bold. $^\star $ means the method is based on a pre-trained model. $^{\romannumeral1}$ means Score-DD based on VE SDE; $^{\romannumeral2}$ means Score-DD based on VP SDE; $^{\romannumeral3}$ means Score-DD based on sub-VP SDE.}
\label{table1}
\end{table*}

\section{Experiment}
\label{sectionD}

\subsection{Datasets}

We conducted tests on common benchmarks to validate the effectiveness of our proposed approach, Score-DD. We describe in detail the data sets used. \textbf{MVTec AD dataset}~\cite{PaulBergmann2019MVTecA} contains 5354 high-resolution images, which is specifically utilized in the unsupervised VDD task. It contains 10 objects and 5 texture categories, and each category contains 60-320 training samples and about 100 test samples. \textbf{BeanTech AD dataset}~\cite{9576231} is an industrial dataset containing 2540 high-resolution images of three products. \textbf{MNIST}~\cite{MNIST} contains 60k training and 10k test $28\times28$ gray-scale handwritten digit images.

% \noindent \textbf {Dataset} The MVTec dataset, which contains 10 objects and 5 texture categories, is specifically utilized in the field of unsupervised anomaly detection. Images in every category are high-resolution images and have already been divided into training dataset and testing dataset. Specially, training dataset only includes anomaly-free images and testing dataset includes both normal and anomaly images that containing various types of natural defects. Meanwhile, every anomaly test image is accompanied with pixel-level annotations of defects.

\subsection{Experiment setup} 

% According to the convention in the previous work, all images in MVTec are resized to a specific size and we perform anomaly detection and location task on one category at a time. We take area under the receiver operating characteristic curve (AUROC) the evaluation metric for both anomaly detection and localization. 

All of the images in the aforementioned dataset are resized to $256\times256$ pixels, except for MNIST which is resized to $32\times32$ pixels. We train a score-based model based on NCSN++ and set 2000 diffusion timesteps for MVTec AD and BTAD, and we train a score model based on DDPM++ and set 1000 timesteps for MNIST. We use the area under the receiver operating characteristic curve (AUROC) the evaluation metric for both defect detection and localization. In the inference stage, after getting the anomaly map, we leverage it to evaluate the AUROC metric for the location task and the maximum value of each anomaly map to evaluate the AUROC metric for the detection task. Other experiment details are presented in {Appendix B}.

%   For VE SDE, we select specific serial number of feature maps according to the visual effect and use a simple linear scheme to select $\left \{ t\right \}$ set that get one $t$ point every 50 steps from one maximum value, and iterate to one minimum value to form the final $\left \{ t\right \}$ set.
% Notably, about carpet class, we perform a simple normalization of the feature map by subtracting its mean and dividing by the variance, keeping the positive data, mainly because the feature maps at different initial moments $t$ has much noise, as Fig.~\ref{T Map} shows. While the other classes of defect maps at different $t$ show different parts of defects which are complementary to each other.

% : \textbf{1)} uses FIR upsampling/downsampling, \textbf{2)} rescales skip connections, \textbf{3)} employs BigGAN-type residual blocks, \textbf{4)} uses 2 residual blocks per resolution, and \textbf{5)} uses "input skip" for input and "out skip" for output.

\subsection{State-of-the-Art Comparison} 

\textbf{Results on MVTec AD} are shown in \cref{table1}, where \textbf{Score-DD} based on VE SDE and Flow ODE sampling method achieves SOTA \textbf{98.2 image-level AUC} and \textbf{97.8 pixel-level AUC}, Score-DD based on VP SDE and reverse SDE sampling method achieves {96.4 image-level AUC} and {97.4 pixel-level AUC}, and {Score-DD} based on sub-VP SDE and the Flow ODE sampling method achieves {97.5 image-level AUC} and {97.2 pixel-level AUC}. We compare our results with the those of SOTA unsupervised VDD methods on the MVTec AD dataset. Specifically, Score-DD outperforms the AE-based method, RIAD~\cite{VitjanZavrtanik2021ReconstructionBI}, and is only $0.1\%\downarrow$ than the GAN combined with pseudo-defects method, OCR-GAN~\cite{YufeiLiang2022OmnifrequencyCR}. Although our approach still lags behind CFLOW~\cite{DenisAGudovskiy2021CFLOWADRU} and FastFlow~\cite{JiaweiYu12022FastFlowUA}, works combined Flow with pre-trained models, as well as some others based on pre-trained models, like PatchCore~\cite{Roth_2022_CVPR}, Score-DD does not depend on pre-trained models that contain rich semantic information and have some comparability in some classes. For the methods of creating pseudo-defects to transform unsupervised learning into supervised learning, Score-DD also outperform CutPaste~\cite{ChunLiangLi2021CutPasteSL}, while DR{\AE}M~\cite{VitjanZavrtanik2021DRAEMA}, which uses additional data to create defects and specifically designed reconstruction model and anomaly segmentation model for VDD, achieved SOTA results, Score-DD also outperforms it by $0.2\%\uparrow$ in the detection task and $0.5\%\uparrow$ in the localization task.

\begin{table}[h]\footnotesize
  \centering
\begin{tabular}{cccccc} 
\hline
Class & Panda$^\star$ & PaDiM$^\star$ & FastFlow$^\star$ & VT-ADL & OURS       \\ 
\hline
1     & 96.4/96.4     & 99.4/97.2     & -/95     & -/99   & 99.2/97.7  \\
2     & 81.0/94.1     & 79.5/95.2     & -/96     & -/94   & 81.1/95.2  \\
3     & 69.8/98.0     & 99.4/98.7     & -/99     & -/77   & 99.1/98.3  \\
Mean  & 82.4/96.2     & 92.7/97.0     & -/97     & -/79   & 93.1/97.1  \\
\hline
\end{tabular}
  \caption{Defect detection (\textbf{left}) and localization (\textbf{right}) performance on BTAD dataset.}
  \label{table2}
\end{table}

\textbf{Results on BeanTech AD} are shown in \cref{table2}, where Score-DD based on VP SDE and the Flow ODE achieves \textbf{93.4 image-level AUC} and \textbf{97.1 pixel-level AUC}. Ours outperforms the reconstruction-based method depended on the transformation model, VT-ADL~\cite{9576231}. And compared with the methods relied on pre-trained models, including Panda~\cite{Reiss_2021_CVPR}, PaDiM~\cite{defard2021padim}, and FastFlow~\cite{JiaweiYu12022FastFlowUA}, we achieve new SOTA results.

\begin{table}[h]\scriptsize
\begin{tabular}{c|ccccc}
\hline
Method & ARAE  & OCSVM & AnoGAN & DSVDD & CapsNet$_{pp}$ \\
AUROC  & 97.5  & 96.0  & 91.4   & 94.8  & 97.7    \\ \hline
Method & OCGAN & LSA   & U-Std$^\star$  & MKDAD$^\star$ & OURS    \\
AUROC  & 97.5  & 97.5  & 99.35  & 98.71 & 95.44   \\ \hline
\end{tabular}
  \caption{Quantitative results of AUROC on the MNIST dataset}
  \label{table3}
\end{table}

\textbf{Results on MNIST} are displayed in \cref{table3}. We want to value Score-DD performance on a large dataset, which achieves a \textbf{95.44 image-level AUC}. Compared with the methods relied on pre-trained models, U-Std~\cite{Bergmann_2020_CVPR} and MKDAD~\cite{Salehi_2021_CVPR}, we are about $3.91\%\downarrow$ below the best result. More fairly, compared with unsupervised methods, including ARAE~\cite{SALEHI2021726}, OCSVM~\cite{958946}, AnoGAN~\cite{2018arXiv180904758L}, DSVDD~\cite{pmlr-v80-ruff18a}, CapsNet$_{pp}$~\cite{10.1007/978-3-030-47358-7_39}, OCGAN~\cite{Perera_2019_CVPR} and LSA~\cite{Abati_2019_CVPR}. We come in $2.3\%\downarrow$ below the top score.

\begin{table}[h]\small
  \centering
\begin{tabular}{cccc} 
\hline
Case  & \begin{tabular}[c]{@{}c@{}}Reconstruction \\Loss\end{tabular} & \begin{tabular}[c]{@{}c@{}}Score-DD \\(w/o $T$ scales)\end{tabular} & \begin{tabular}[c]{@{}c@{}}Score-DD \\(w/ $T$ scales)\end{tabular}  \\ 
\hline
AUROC & 84.85
/89.34
                                                     & 91.30
/96.24
                                                         & 98.24/97.78                                                        \\
\hline
\end{tabular}
  \caption{Quantitative results for ablation studies on MVTec AD.}
  \label{table4}
\end{table}

\subsection{Ablation study and analysis}

We do comparative experiments to confirm the efficacy of submodules. Specifically, we run a continuous version experiment (Score-DD without $T$ scales), iterating from the maximum moment in $\left \{t\right \}$ to near the smallest moment $\epsilon$, and then calculate difference of the $s_w(\cdot,\epsilon)$ and its feature map, to simulate score difference, $||\nabla_{\mathbf{x}'}\mathrm{log}\ p_{t}(\mathbf{x}'(\epsilon))-\nabla_{\mathbf{x}''}\mathrm{log}\ p_{t}(\mathbf{x}''(\epsilon))||^2$, and also do a experiment based on reconstruction loss between $\mathbf{x}'(0)$ and $\mathbf{x}''(0)$ as a metric for VDD, $||\mathbf{x}'(0)-\mathbf{x}''(0)||^2$. When comparing the case with reconstruction loss to Score-DD without $T$ scales in the \cref{table4}, it is clear that the score as a metric is more useful for VDD. Furthermore, when compared to the case without $T$ scales, $T$ scales technique can improve performance. This is because it iterates just a few steps, some normal pixels have little opportunity to access other high probability density regions, alleviating the problems discussed above in \cref{problems} further, while it also leverages different semantic information at different moments.

% \textbf{SDEs and Sampling method} We test different SDEs: VE, VP and sub-VP SDE, which are introduced in \cite{YangSong2020ScoreBasedGM}. We also compare results from iterating reverse SDE and the Flow ODE. Each SDE has different discrimination ability for different classes. From Table. \ref{table2}, in defect localization, VP SDE performs better, while in defect classification task, sub-VP SDE performs better. Plus, the results with the Flow ODE and reverse SDE have similar results.

\subsection{Computational complexity}

We propose $T$ scales technique to speed up the inference process. Take an example from the real case in the MVTec AD dataset. After we train a score model that needs $S=2000$ iteration steps to generate images, we set inital timesteps as $t=250/2000$; Thus, the inference-time efficiency of Score-DD without $T$ scales is $O(t*S-1+2)=O(251)$. However, with $T$ scales, we develop a set containing different timesteps $\{t\}$. For example, we can take a $t$ from 250 to 50 every 50 steps, $\{t\} = \{250,200,150,100,50\}/2000$ containing $T=5$ different timesteps. Therefore, the minimal inference-time efficiency is $O(T*(r+2))=O(T*3)=O(15)$, which is more efficient than AnoDDPM~\cite{Wyatt_2022_CVPR} whose inference-time efficiency is $O(t*S)=O(250)$. The inference time efficiency we calculate is based on the number of runs of the neural network. What's more, $T$ scales technique also makes the sequential iteration split into parallel cases, and each $t$ case needs to iterate just a few $r$ steps. Therefore, it can also run in parallel to speed things up even further. Moreover, the community is also exploring some ways to accelerate the score model and diffusion model.

\subsection{Limitation and future work}

Because our work is based on the assumption that data located in the low density region of the normal data distribution are considered abnormal, our work is restricted to the unsupervised VDD setting, and does not win comparative edges in some AD settings. We think part of the reason is that in the AD setting, some changes that are significantly different from the training data, such as changes in the complex background, are of more interest to the model than information about the categories and objects. So it is better to have additional category information or classifiers for the auxiliary~\cite{2022arXiv220304306W}. Besides, our approach is sensitive to feature maps that are fixedly selected and cannot be adaptively adjusted according to different images. Thus, difficulties are encountered in experimentally tuning a large dataset like MNIST, resulting in some performance degradation. However, because the training process and loss function don't change, this can be alleviated by easily extending our method with a professional and generalized classification or segmentation network with self-supervised (e.g., pseudo-defects) or semi-supervised (i.e., several available abnormal samples with image or pixel labels) methods. Since the purpose of this paper is to explore unsupervised solutions that are applicable to the score model and do not depend on other models or mechanisms, we leave this work for the future.

\section{Conclusion}
\label{section:conclusion}

We propose to use a score-based generative model for unsupervised VDD. Our research indicates that employing a metric for VDD that is linked to the probability density of normal data, e.g., a score value, can efficiently handle the challenge of reconstructed images that differ from the original images with normal pixels in pixel space. In addition, we propose to use $T$ scales to solve the problem of slow speed due to the need of iterating multiple steps, and since we only need to iterate a few steps at each $t$ moment, it doesn't deviate the normal pixels too much from the original data, which in turn improves the accuracy. Without using additional data, algorithms, and models, we have achieved competitive performance on several datasets. 

%----------------------------------------------------------------------------------------------------------------------------------------------------------------------------------------------------------------------------------------------------------------------------------------------------------------------------------------------------------------------------------------------------------------------------------------

%%%%%%%%% REFERENCES
{\small
\bibliographystyle{ieee_fullname}
\bibliography{egbib}
}

\clearpage
\appendix
\section{Supplementary formula}
\subsection{Proof for the true trajectory}
\label{A}

First, we define the true trajectory from $\mathbf{x}(t)$ to $\mathbf{x}(0)$ in the sense that after obtaining $\mathbf{x}(t)$ by injecting noise into $\mathbf{x}(0)$, the path is iterated back to the original $\mathbf{x}(0)$ through reverse stochastic process. The proof of the key formulation about the true trajectory from $\mathbf{x}(t)$ to $\mathbf{x}(0)$ is given below.

% Besides, we will talk about the difference of $\textbf{whole-score}$ and $\textbf{self-score}$.

The objective of a generative model is to generate samples that satisfy the distribution of the given training data. Recalling Section 3.1, we train a neural network to fit the score function: $\nabla_{\mathbf{x}}\mathrm{log}\ p_{t}(\mathbf{x}(t))$ of given dataset $\mathbf{X}_N$, which guarantees the score-based generative model eventually generate $\mathbf{x}(0)\sim p_{0}(\mathbf{x})$ through reverse process, where $p_{0}(\mathbf{x})\approx p_{data}(\mathbf{x})$ by definition. Here,

% \begin{equation}
\begin{align}
&\frac{\partial \mathrm{log}\ p_{t}(\mathbf{x}(t))}{\partial \mathbf{x}(t)}=\frac{1}{p_t(\mathbf{x}(t))}\frac{\partial p_t(\mathbf{x}(t))}{\partial \mathbf{x}(t)}\notag\\
&=\frac{1}{p_t(\mathbf{x}(t))}\frac{\partial }{\partial \mathbf{x}(t)}\int p_0(\mathbf{x}(0))p_{0t}(\mathbf{x}(t)|\mathbf{x}(0))\mathrm{d}\mathbf{x}(0)\notag\\
&=\frac{1}{p_t(\mathbf{x}(t))}\int p_0(\mathbf{x}(0))\frac{\partial p_{0t}(\mathbf{x}(t)|\mathbf{x}(0))}{\partial \mathbf{x}(t)}\mathrm{d}\mathbf{x}(0)\notag\\
&=\frac{1}{p_t(\mathbf{x}(t))}\int p_0(\mathbf{x}(0))p_{0t}(\mathbf{x}(t)|\mathbf{x}(0))\frac{\partial \mathrm{log}\ p_{0t}(\mathbf{x}(t)|\mathbf{x}(0))}{\partial \mathbf{x}(t)}\mathrm{d}\mathbf{x}(0) \notag\\
&=\int \frac{p_0(\mathbf{x}(0))p_{0t}(\mathbf{x}(t)|\mathbf{x}(0))}{p_t(\mathbf{x}(t))}\frac{\partial \mathrm{log}\ p_{0t}(\mathbf{x}(t)|\mathbf{x}(0))}{\partial \mathbf{x}(t)}\mathrm{d}\mathbf{x}(0) \tag{6}. \label{score}
\end{align}
% \end{equation}

However, regarding the true trajectory from $\mathbf{x}(t)$ to $\mathbf{x}(0)$, we can consider that
$p_{data}(\mathbf{x})$ degenerates to a one-point distribution with mean $\mathbf{x}(0)$ and variance 0, denoted as $p_{data}''(\mathbf{x})$:

\begin{equation}\tag{7}
p_{data}''(\mathbf{x}) =\begin{cases}

1, & \mathbf{x}=\mathbf{x}(0)\\
0, & \textup{Others}

\end{cases}
% \nonumber
\label{eq9}
\end{equation}

Therefore, $\nabla_{\mathbf{x}}\mathrm{log}\ p_{t}''(\mathbf{x}(t))=\nabla_{\mathbf{x}}\mathrm{log}\ p_{0t}''(\mathbf{x}(t)|\mathbf{x}(0))$. As discussed in Section 3.1, if the drift coefficient $f(t)$ of the SDE is linear, the transition density is Gaussian $p_{0t}(\mathbf{x}(t)|\mathbf{x}(0))=\mathcal{N}(\mathbf{x}(t);\mu(t)\mathbf{x}(0), \sigma(t)^2\textup{I})$. Thus, $\nabla_{\mathbf{x}}\mathrm{log}\ p_{t}''(\mathbf{x}(t))=-\frac{\mathbf{x}(t)-\mu(t)\mathbf{x}(0)}{\sigma(t)^2}=-\frac{\mathbf{z}(t)}{\sigma(t)}$, where $\mathbf{x}(t)=\mu(t)\mathbf{x}(0)+\sigma(t)\mathbf{z}(t)$, $\mathbf{z}(t)\sim\mathcal{N}(0,\textup{I})$, denoted as self-score $s_e(\cdot,t)$ in our paper. Therefore, we can obtain Eq.(6) and Eq.(7) by plugging $\nabla_{\mathbf{x}}\mathrm{log}\ p_{t}''(\mathbf{x}(t))$ into Eq.(2) and Eq.(5) respectively.

% \begin{equation}
\begin{align}
    \mathrm{d}\mathbf{x}(t) &= (f(t)\mathbf{x}(t)-g(t)^{2}(-\frac{\mathbf{z}(t)}{\mathbf{\sigma}(t)}))\mathrm{d}\bar{t}+g(t)\mathrm{d\mathbf{\bar{w}}}(t).\tag{8}\label{Se1}\\
    \mathrm{d}\mathbf{x}(t) &= (f(t)\mathbf{x}(t)-\frac{1}{2}g(t)^{2}(-\frac{\mathbf{z}(t)}{\mathbf{\sigma}(t)}))\mathrm{d}t. \tag{9}\label{Se2}
\\
\nonumber
\end{align}
% \end{equation}

The path obtained by iterating \cref{Se1} is represented as the true trajectory from $\mathbf{x}(t)$ to $\mathbf{x}(0)$ with the Reverse SDE in Eq.(2), and the path from \cref{Se2} as the true trajectory with the probability flow ODE in Eq.(5). In addition, when training the score-based model, whole-score is actually evaluated through self-score $-\frac{\mathbf{z}(t)}{\sigma(t)}$ of each sample in the training dataset. And we can provide more insights of the principle of \textbf{score-AD} by analyzing the difference of whole-score and self-score in calculation ways with \cref{score}.

\subsection{Details VE, VP and sub-VP SDEs}
\label{B}

We follow the definitions of VE, VP and sub-VP SDEs as in~\cite{YangSong2020ScoreBasedGM}, and show them in order:

\begin{equation}\tag{10}
\begin{cases}

\mathrm{d}\mathbf{x}(t) = \sqrt{\frac{\mathrm{d}[\sigma ^{2}(t)]}{\mathrm{d}t}}\mathrm{d}\mathbf{w}(t),  \\
\mathrm{d}\mathbf{x}(t) = -\frac{1}{2}\beta (t)\mathbf{x}(t)\mathrm{d}t+\sqrt{\beta (t)}   \mathrm{d}\mathbf{w}(t),  \\
\mathrm{d}\mathbf{x}(t) = -\frac{1}{2}\beta (t)\mathbf{x}(t)\mathrm{d}t+\sqrt{\beta (t)(1-e^{-2\int_0^{t}\beta (s)\mathrm{d}s})}\mathrm{d}\mathbf{w}(t). \\

\end{cases}
% \nonumber
\end{equation}

VE SDE refers to Variance Exploding (VE) SDE because VE SDE always gives a process with exploding variance when t increases. The Variance Preserving (VP) SDE yields a process with a fixed variance of one when the initial distribution has unit variance. The variance of the stochastic process induced by the sub-VP SDE is always bounded by the VP SDE at every intermediate time step. See ~\cite{YangSong2020ScoreBasedGM} for more information.

Because VE, VP and sub-VP SDEs all have linear drift coefficients $f(t)$, their corresponding transition densities $p_{0t}(\mathbf{x}(t)|\mathbf{x}(0))$ are all Gaussian:
% p_{0t}(\mathbf{x}(t)|\mathbf{x}(0)) = 
\begin{equation}\tag{11}
\begin{cases}

\mathcal{N}(\mathbf{x}(t);\mathbf{x}(0),[\sigma ^{2}(t)-\sigma ^{2}(0)]\textup{I}) & \textup{(VE SDE)}\\
\mathcal{N}(\mathbf{x}(t);\mathbf{x}(0)e^{-\frac{1}{2}\int_{0}^{t}\beta (s)\mathrm{d}s},\textup{I}-\textup{I}e^{-\int_{0}^{t}\beta (s)\mathrm{d}s}) & \textup{(VP SDE)}\\
\mathcal{N}(\mathbf{x}(t);\mathbf{x}(0)e^{-\frac{1}{2}\int_{0}^{t}\beta (s)\mathrm{d}s},[1-e^{-\int_{0}^{t}\beta (s)\mathrm{d}s}]^{2}\textup{I}) & \textup{(sub-VP SDE)}

\end{cases}.
% \nonumber
\end{equation}

In detail, there are discretizations of SDEs. For VE SDE, set 

\begin{equation}\tag{12}
\sigma(t)=\begin{cases}

\sigma_{\textup{min}}(\frac{\sigma_{\textup{max}}}{\sigma_{\textup{min}}})^{t}, & t\in(0,1]\\
0, &t=0.

\end{cases}
\end{equation}

For both VP SDE and sub-VP SDE, they are set as:

\begin{equation}\tag{13}
\beta(t)=\beta_{\textup{min}}+t(\beta_{\textup{max}}-\beta_{\textup{min}}).
\end{equation}

\subsection{Algorithm}
\label{C}

From \cref{Se1} and \cref{Se2}, we can approach the original $\mathbf{x}(0)$ from $\mathbf{x}(t)$ if we know $\mathbf{z}(t)$ relative to $\mathbf{x}(0)$ at each moment. 
% Therefore, iterating over Eq.(\ref{RSDE with self-score.}) or Eq.(\ref{DXTP}) can eventually reach the vicinity of $\mathbf{x}(0)$. 
It should be noted that the $\mathbf{z}(t_i)$ changes constantly over steps, and by deforming Eq.(4), $\mathbf{z}(t)=\frac{\mathbf{x}(t)-\mu(t)\mathbf{x}(0)}{\sigma(t)}$, we can know it will be updated along with $\mathbf{x}''(t_i)$. At the current step, as the $\mathbf{z}(t_i)$ is known in advance, we can bring self-score computed by $\mathbf{z}(t_i)$ to \cref{Se1} or \cref{Se2} to obtain $\mathbf{x}''(t_{i+1})$. Through above deformation of Eq.(4) and $\mathbf{x}''(t_{i+1})$, we can therefore obtain $\mathbf{z}(t_{i+1})$. Repeating this process, we can obtain the complete trajectory of $\mathbf{{x}}(t)$ to $\mathbf{{x}''}(t_r)$, or ultimately to $\mathbf{{x}''}(0)\approx\mathbf{{x}}(0)$. Therefore, after $r$ steps, we can feed the sets $\{x'(t_r)\}$ and $\{x''(t_r)\}$ (each of which has a capacity of $T$) into the score-based model. We conclude the algorithm about \textbf{Score-AD}. Algorithms 1 and 2 denote the reverse diffusion process with the probability flow ODE and Reverse SDE separately. $\mathbf{x}(0)\in\mathbf{X}_{N+A}$ is a test image, $\{t\}$ is a set of different initial times with capacity of $T$, and $r$ is the number of iteration steps.

% \begin{algorithm}[H]
    
%     \caption{Score AD with Reverse SDE}
%     \label{fcrsde}
%     \begin{algorithmic}[1]
%     \Require $\mathbf{x}_{m};$
%     $\Psi=\left\{\mathbft_i \right\}_{i=1}^{n};$ S;
%     % \Ensure Anomaly Maps
%     \For {$\mathbf{all}\ (\mathbft_i )\in \Psi$}
%         \State ${\mathbf{x}(t) = \mathbf{\mu} (t )\mathbf{x}_m + \mathbf{\sigma}(t)\mathbf{z}(t)}$
%         \State $\mathbf{z}(t)\sim \mathcal{N}(0,I)$
%             \For {$ s = 0\ \textbf{to}\ \textup{S}-1 $}
%             \State \begin{aligned}
%             \mathbf{{x}'}(t_i) &= \mathbf{{x}'}(t_i) - (f(t-s)\mathbf{{x}'}(t_i)-g(t_i)^{2} s_{\theta } (\mathbf{{x}'},t_i))+\mathbf{n}(t_i)
%             \\ 
%             \mathbf{{x}''}(t_i) &= \mathbf{{x}''}(t_i)-(f(t-s)\mathbf{{x}''}(t-s)-g(t_i)^{2} (-\frac{\mathbf{z}(t_i)}{\sigma({t-s)}}))+\mathbf{n}(t_i)
%             \\
%             \mathbf{z}(t_i)&=\frac{\mathbf{{x}''} (t_i)-\mathbf{\mu}(t-s-1){\mathbf{x}}_m}{\sigma(t_i)}
%             \\
%             \mathbf{n}(t_i)&\sim \mathcal{N}(0,I)
%             \end{aligned}
%             \EndFor
%             \State Input $\mathbf{{x}'}(t-\textup{S})$ and $\mathbf{{x}''}(t-\textup{S})$ to the score model
%             \EndFor
%         \State Add or multiply feature maps\\
%         \Return Anomaly Map
    
%     % \Until {Stop condition reached} \

%     \end{algorithmic}
% \end{algorithm}

% \section{Implementation Details}
% % \subsection{Implementation Details}
% \label{D}

% The problems discussed in Section 4 still exist, and we just select images with good visual effect.

% 不太好动，浮动模式设置成t就会消失，最好自己手动调整固定位置
% \columnseprule=2pt   
\setlength{\columnsep}{0.5cm}

\floatname{algorithm}{Algorithm}
\renewcommand{\algorithmicrequire}{\textbf{Require:}}
\renewcommand{\algorithmicensure}{\textbf{Output:}}
% \begin{multicols}{2}    
\begin{algorithm}[t]\small

    \caption{Score-AD with the flow ODE}
    \label{f1}
    \begin{algorithmic}[1]
    
    \Require $\mathbf{x}(0)$;$\{t\}$; $r$;
    % \Ensure Anomaly Maps
    
    \For {$t\in \{t\}$}
        \State $\mathbf{x}(t) = \mathbf{\mu} (t )\mathbf{x}(0)$
        \State $\mathbf{z}(t)\sim \mathcal{N}(0,I)$
        \State $\mathbf{x}(t_0) = \mathbf{x}(t)+ \mathbf{\sigma}(t)\mathbf{z}(t)$
            \For {$ i=0\ \textbf{to}\ r-1 $}
        \State 
            \begin{math}
            \begin{aligned}
            \Delta t&=t_i-t_{i+1}\\
            \\
            \mathbf{{x}'}(t_{i+1}) &= \mathbf{{x}'}(t_{i}) - f(t_{i})\mathbf{{x}'}(t_{i})\Delta t\\
            \mathbf{{x}'}(t_{i+1}) &= \mathbf{{x}'}(t_{i+1}) +\frac{1}{2} g(t_{i})^{2} s_{\theta } (\mathbf{{x}'},t_{i})\Delta t~\\
            ~\\
            \mathbf{{x}''}(t_{i+1}) &= \mathbf{{x}''}(t_i)-f(t_i)\mathbf{{x}''}(t_i)\Delta t\\
            \mathbf{{x}''}(t_{i+1}) &=\mathbf{{x}''}(t_{i+1}) +\frac{1}{2}g(t_{i})^{2}
            (-\frac{\mathbf{z}(t_i)}{\sigma({t_i)}})\Delta t~\\
            ~\\
            \mathbf{z}(t_{i+1})&=\frac{\mathbf{{x}''} (t_{i+1})-\mathbf{\mu}(t_{i+1}){\mathbf{x}}(0)}{\sigma(t_{i+1})}
            \end{aligned}
            \end{math}
            \EndFor
        \State Input $\mathbf{{x}'}(t_r)$ and $\mathbf{{x}''}(t_r)$ to the score model
        \EndFor
    \State Add or multiply feature maps
    \\
    \Return Anomaly Map
    
    % \Until {Stop condition reached} \
    
    \label{fcfode}
    \end{algorithmic}
    
\end{algorithm}

\begin{algorithm}[t]\small
    
    \caption{Score-AD with Reverse SDE}
    \begin{algorithmic}[2]
    \label{f2}
    \Require $\mathbf{x}(0)$;$\{t\}$; $r$;
    % \Ensure Anomaly Maps
    
    \For {$t\in \{t\}$}
        \State $\mathbf{x}(t) = \mathbf{\mu} (t )\mathbf{x}(0)$
        \State $\mathbf{z}(t)\sim \mathcal{N}(0,I)$
        \State $\mathbf{x}(t_0) = \mathbf{x}(t)+ \mathbf{\sigma}(t)\mathbf{z}(t)$
            \For {$ i=0\ \textbf{to}\ r-1 $}
        \State 
            \begin{math}
            \begin{aligned}
            \Delta t&=t_i-t_{i+1}\\
            \mathbf{n}(t_i)&\sim \mathcal{N}(0,I)\\
            \mathbf{{x}'}(t_{i+1}) &= \mathbf{{x}'}(t_{i}) - f(t_{i})\mathbf{{x}'}(t_{i})\Delta t\\
            \mathbf{{x}'}(t_{i+1}) &= \mathbf{{x}'}(t_{i+1})+g(t_{i})^{2} s_{\theta } (\mathbf{{x}'},t_{i})\Delta t\\ 
            \mathbf{{x}'}(t_{i+1}) &= \mathbf{{x}'}(t_{i+1})+g(t_{i})\sqrt{\Delta t}\mathbf{n}(t_i)\\
            \mathbf{{x}''}(t_{i+1}) &= \mathbf{{x}''}(t_i)-f(t_i)\mathbf{{x}''}(t_i)\Delta t\\
            \mathbf{{x}''}(t_{i+1}) &=\mathbf{{x}''}(t_{i+1}) +g(t_{i})^{2} (-\frac{\mathbf{z}(t_i)}{\sigma({t_i)}})\Delta t\\
            \mathbf{{x}''}(t_{i+1}) &=\mathbf{{x}''}(t_{i+1}) +g(t_{i}) \sqrt{\Delta t}\mathbf{n}(t_i)\\            
            \mathbf{z}(t_{i+1})&=\frac{\mathbf{{x}''} (t_{i+1})-\mathbf{\mu}(t_{i+1}){\mathbf{x}}(0)}{\sigma(t_{i+1})}
            \end{aligned}
            \end{math}
            \EndFor
        \State Input $\mathbf{{x}'}(t_r)$ and $\mathbf{{x}''}(t_r)$ to the score model
        \EndFor
    \State Add or multiply feature maps
    \\
    \Return Anomaly Map
    
    % \Until {Stop condition reached} \

    \end{algorithmic}
    
\end{algorithm}

% \end{multicols}

\section{Implementation Details}
% \subsection{Implementation Details}
\label{D}

Below, we add additional implementation details for each experiment.

\textbf{MNIST Figure.} For Fig. 1, to demonstrate our method and show more insights, we train a score model on a subset with category ``1" on the MNIST dataset while selecting an image with category ``7" for testing. The score model is based on VE SDE, which adopts the U-net architecture and code can be found in the COLAB tutorial of~\url{https://github.com/yang-song/score_sde_pytorch}. We choose $\sigma(t)=(25)^{t}$, set diffusion timesteps as 1000 and initial moment $t=0.2$ to get the final reconstructed image. 

\textbf{MNIST Experiment.} We choose VP SDE. Specially, $\beta_{min}=0.1$, $\beta_{max}=20$, and set diffusion timesteps as 1000. Based on the previous work, we adopt the positional embeddings, the layers in~\cite{NEURIPS2020_4c5bcfec} to condition the score model on continuous time variables. As for architecture of score-based model, we take DDPM++ structure introduced in ~\cite{YangSong2020ScoreBasedGM}:  \textbf{1)} rescales skip connections; \textbf{3)} employs BigGAN-type residual blocks; \textbf{4)} uses 2 residual blocks per resolution; and \textbf{5)} uses "residual" for input. Please see~\cite{YangSong2020ScoreBasedGM} and~\href{https://github.com/yang-song/score_sde_pytorch}{yang-song/score\_sde\_pytorch} to get more information.

\textbf{Exploratory experiment.} For Fig.~3, based on the instantiation scheme of VE SDE, we choose $\sigma_{\textup{min}}=0.1$ and $\sigma_{\textup{max}}=20$. Specially, we select three data points $(-6.0,5.0), (5.17,5.2), (-4.2,-4.3)$. Based on our assumption and normal data distribution, $(-6.0,5.0)$ is anomaly data. Consistent with the results in Fig.~3, the difference in whole-score value between the ``reconstructed and original noisy'' data pairs is much larger than for normal data.

\textbf{MvTec AD and BeanTech AD dataset.} The MVTec AD dataset is available at~\url{https://www.mvtec.com/company/research/datasets/mvtec-ad/} and the BTAD dataset is available at~\url{https://github.com/pankajmishra000/VT-ADL}. For VE SDE, we choose $\sigma_{\textup{min}}=0.01$ and $\sigma_{\textup{max}}=348$. For VP SDE and sub-VP SDE, we select $\beta_{\textup{min}}=0.1$ and $\beta_{\textup{max}}=20$. Based on the previous work, we use random Fourier feature embeddings layers introduced in~\cite{NEURIPS2020_55053683} to condition the score model on continuous time variables for VE SDE, and the scale parameter of Fourier feature embeddings is fixed at 16. For VP and sub-VP SDE, we adopt positional embeddings. As for the architecture of a score-based model, we take the NCSN++ structure for all SDEs : \textbf{1)} uses FIR upsampling/downsampling; \textbf{2)} rescales skip connections; \textbf{3)} employs BigGAN-type residual blocks; \textbf{4)} uses 2 residual blocks per resolution; and \textbf{5)} uses "residual" for input and no progressive growing architecture for output.
The code for the score-based model can be found at~\href{https://github.com/yang-song/score_sde_pytorch}{yang-song/score\_sde\_pytorch}. The results of Tab. 2 are based on VPSDE and Flow ODE sampling methods. For the results of Tab. 1, to choose a set of different initial moment $\{t\}$, we adjust the maximum and minimum $t$ in ${\{t\}}$, and then take time stamp every 50 steps interval. The selected feature maps, or $\{t\}$ set, work well and are effective in most cases, but not on every image. Therefore, an adaptive feature map selection strategy would be helpful and could be our future extension.

\end{document}